\documentclass{article}

\usepackage{arxiv}

\usepackage[utf8]{inputenc} % allow utf-8 input
\usepackage[T1]{fontenc}    % use 8-bit T1 fonts
\usepackage{hyperref}       % hyperlinks
\hypersetup{
    colorlinks=true,
    linkcolor=blue,
    filecolor=magenta,      
    urlcolor=cyan,
}
\usepackage{url}            % simple URL typesetting
\usepackage{booktabs}       % professional-quality tables
\usepackage{amsfonts}       % blackboard math symbols
\usepackage{nicefrac}       % compact symbols for 1/2, etc.
\usepackage{microtype}      % microtypography
\usepackage{subcaption}
\usepackage{graphicx}
\usepackage{textcomp, setspace}
\usepackage{amsmath}
\usepackage{authblk}
\title{SwinUnet3D - A hierarchical architecture for deep traffic prediction using shifted window transformers}
\author{Alabi Bojesomo}
\author{Hasan Al Marzouqi}
\author{Panos Liatsis}

\affil{Khalifa University, Abu Dhabi, UAE \authorcr
  \{\tt alabi.bojesomo, hasan.almarzouqi, panos.liatsis\}@ku.ac.ae}
  
\begin{document}

\maketitle

\begin{abstract}
Traffic forecasting is an important element of mobility management, an important key that drives the logistics industry. Over the years, lots of work have been done in Traffic forecasting using time series as well as spatiotemporal dynamic forecasting.  In this paper, we explore the use of vision transformer in a UNet setting. We completely remove all convolution-based building blocks in UNet, while using 3D shifted window transformer in both encoder and decoder branches.  In addition, we experiment with the use of feature mixing just before patch encoding to control the inter-relationship of the feature while avoiding contraction of the depth dimension of our spatiotemporal input. The proposed network is tested on the data provided by Traffic Map Movie Forecasting Challenge 2021(Traffic4cast2021), held in the competition track of Neural Information Processing Systems (NeurIPS). Traffic4cast2021 task is to predict an hour (6 frames) of traffic conditions (volume and average speed)from one hour of given traffic state (12 frames averaged in 5 minutes time span).  Source code is available online at \href{href: code}{\textit{https://github.com/bojesomo/Traffic4Cast2021-SwinUNet3D}}.
\end{abstract}



\section{Introduction}
Traffic forecasting is important for managing traffic flow in cities around the world. For instance, knowledge of the traffic situation can help reduce congestion while it also increases logistic efficiency. Deep learning methods used for traffic forecasting depends on the presentation of the data. When the dynamically changing traffic situation is presented in form of a Map, Graph Neural Network (GNN) is majorly used with profound performance improvement over time \cite{pmlr-v123-martin20a, roy2021unified}. Recently, traffic forecasting task is being presented in spatiotemporal format (video prediction) which enables the use of models applicable to dense prediction. Prior works shows that convolutional neural network (CNN) and long short-time memory can be applied \cite{xu2020good, yu2019crevnet, bojesomo2020traffic, choi2019traffic, choi2020utilizing}. Many researchers use UNet architecture because of the demonstrated high level of performance when solving such problems.  \cite{t4c2020-kopp21a, choi2020utilizing, bojesomo2020traffic, xu2020good}. 

It has been shown that while CNN has been widely used, transformers originating from natural language processing (NLP) not only compete but also lead to stat-of-the-art models in image classification , object detection, semantic segmentation and spatiotemporal forecasting \cite{Dosovitskiy2021_image_transformer, Wang2021PyramidVT, ranftl2021vision_densePrediction,dai2021coatnet}. Following the work of \textit{Dosovitskiy et. al.} \cite{Dosovitskiy2021_image_transformer}, many research problems in computer visoin have been solved using transformers, which directly apply a modified version of the transformer network using patch-based image encoding. 
To circumvent the otherwise computational bottleneck of full attention used in vision transformer, a variety of proposed methods utilize techniques that limit the scope of attention units as used in Swin-Transformer \cite{Liu2021_swin_transformer}, CrossFormer \cite{wang2021crossformer}, CSWin \cite{dong2021cswin} among others. In a contrast to these attempts, \cite{dai2021coatnet} proposed using convolutions in earlier stages while using full attention in later stage of the architecture.  
This architecture named CoAtNet achived  impressive performance in the ImageNet classification problem, ushering us to possibly next paradigm of architectures \cite{dai2021coatnet}.
% Recent state-of-the-art models in ImageNet classification task combines convolution neural network with transformer, ushering us to possibly next paradigm of architectures \cite{dai2021coatnet}. This leads to some more insight on understanding the important of the permutation invariance of transformer owing to the global receptive field, as well as translation equivalence of the local receptive field in convolution \cite{dai2021coatnet}. 
Many pioneering works have used vision transformers to replace the CNN backbone in dense prediction task , including semantic segmentation, and object detection \cite{ranftl2021vision_densePrediction, Wang2021PyramidVT, Dosovitskiy2021_image_transformer, Liu2021_swin_transformer}.
Such decision can be clearly attributed to how attention based network revolutionized natural language processing (NLP) \cite{liu2019roberta, attention_galassi2021, transformer_vaswani2017}.

Transformer usage in vision paradigm starts with the encoding of image patches similar to word embedding in NLP, followed by one-to-one (or modified) use of  transformer network as proposed by \textit{Vaswani et. al.} \cite{transformer_vaswani2017}. Research direction here includes modification of the patch encoding, use of efficient attention networks as well as the inclusion of absolute (or relative) position encoding \cite{zhang2021rest, xie2021segformer, Liu2021_swin_transformer, Liu2021_video_swin_transformer}.

Among promising architecture adopting vision transformer is the shifted window transformer (Swin Transformer) based on its performance in solving several computer vision tasks.
Swin Transformer is unique in its use of local windows which are shifted in subsequent layers, leading to efficient information mixing \cite{Liu2021_swin_transformer}. This architecture has been adopted as backbone for dense prediction. 

Swin Transformer uses a carefully crafted patch information mixing using the shifted window attention technique (hence the name Swin) \cite{Liu2021_swin_transformer}. 
The method resulted in high performing image recognition models using the vision transformer. 
In order to leverage its success in classification, researchers explored the Swin transformer as backbone for dense prediction, object detection and segmentation. A notable use of Swin Transformer for dense prediction can be seen in the work of \textit{Cao et. al.}, where all the blocks of Unet structure are  replaced by Swin transformer blocks \cite{Cao2021_swin_unet}. The modified UNet structure (Swin-UNet) uses patch merging layer for downsampling in the encoder \cite{Wang2021PyramidVT, Cao2021_swin_unet}, as well as patch expanding layers for upsampling in the decoder branch \cite{Cao2021_swin_unet}.

\textit{Liu et. al.} proposed Video Swin transformer, a 3-dimensional (3D) variant of the Swin transformer \cite{Liu2021_video_swin_transformer}. In this context, 3D patch embedding, 3D shifted window multi-head self attention as well as patch merging were proposed to work with 3D inputs. Success of video swin transformer was demonstrated using several challenging video recognition problems and datasets (Kinetics-400, kinetic-600, and Something-Something v2) \cite{Liu2021_video_swin_transformer}.

In this research, we propose a number of improvements in the video Swin transformer \cite{Liu2021_video_swin_transformer}, including 3D patch expanding, and feature mixing layer, as well as a transformer friendly data augmentation process. The proposed network achieved competitive performance without pre-training the network on a large dataset. The proposed architecture and layers are described in detail in Section \ref{sec: methods}, while the experimental results of the proposed solution in traffic4cast2021 challenge are presented in Section \ref{sec: result}.

\section{Methods}
\label{sec: methods}
\subsection{Model Architecture}
Traffic forecasting is sequence-to-sequence task, that can be tackled with transformer models, which were used in NLP with very promising results \cite{liu2019roberta}. The proposed model uses hierarchical spatial reduction in the encoder to capture salient representations of global features. Similar to Swin-UNet \cite{Cao2021_swin_unet}, we use self attention to replace all blocks of a traditional UNet structure (encoder and decoder). 
For merging the skip connected input from the encoder with the main input in the decoder, we propose using; simple addition as used in LinkNet \cite{linknet_Chaurasia2017}; concatenation as used in traditional UNet \cite{unet}; or combination of both. 
Reasoning for the use of simple addition comes from the fact that upsampling in this model is a learned process using patch expanding.
The attention layer used in this research is the shifted window attention proposed in \cite{Liu2021_video_swin_transformer}.

\begin{figure*}[htbp!]
    \centering
    \includegraphics[width=\linewidth]{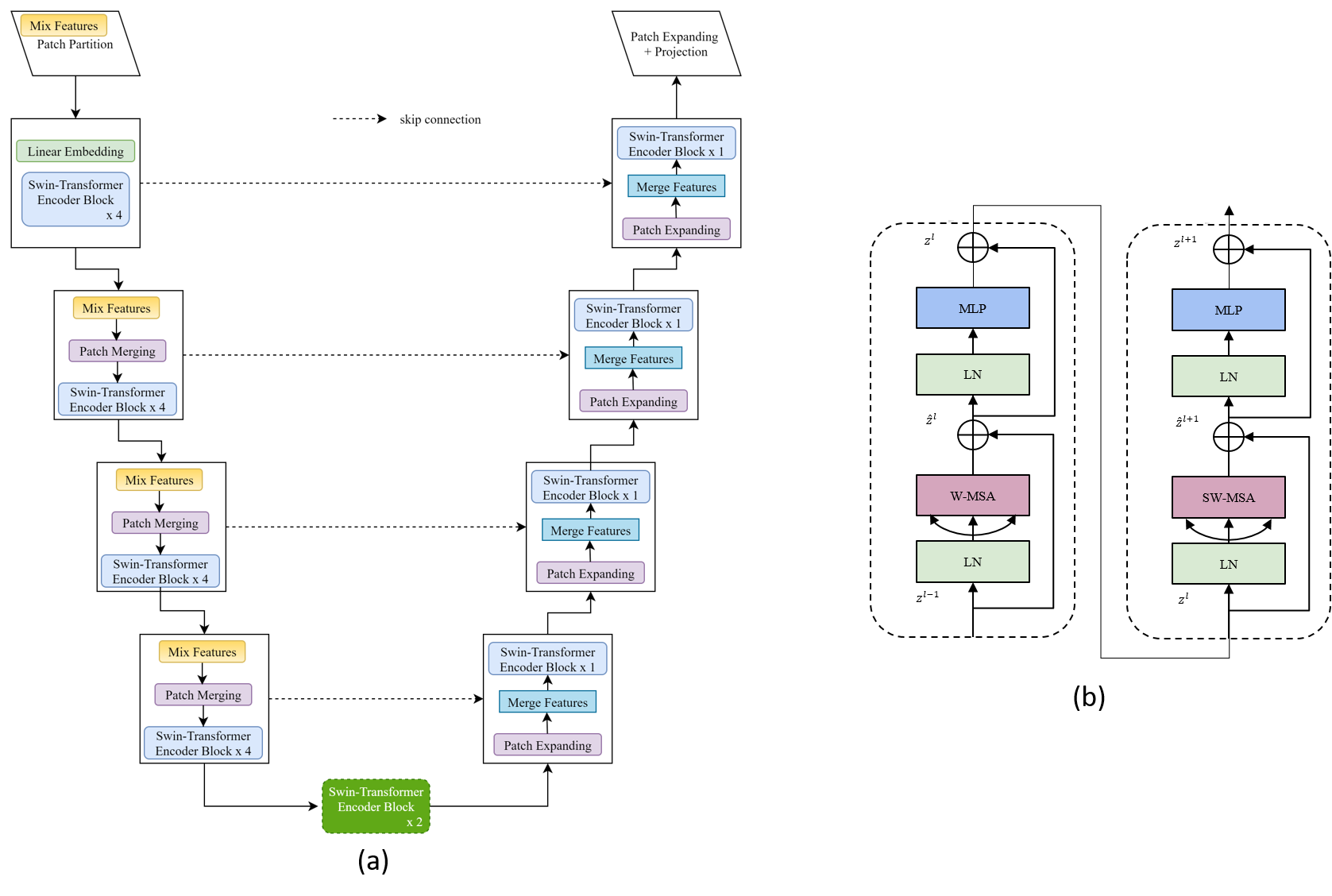}
    \caption{Details of the proposed Spatiotemporal Swin-UNet3D architecture. The network, shown in (a), has four transformer blocks and decoder blocks, respectively. Each of the encoder blocks has a Feature Mixing unit , Patch Merging unit, followed by a number of Swin-Transformers except for the first encoder, which has linear embedding. Also, each of the decoder blocks has a Patch Expanding Unit, Feature Merging Unit, followed by a number of Swin-Transformers. In (b), the stacking of two concurrent self-attention blocks used is illustrated, where windowed attention always follows shifted window attention.}
    \label{fig: architecture}
\end{figure*}

As shown in Fig \ref{fig: architecture}, the input goes through a feature mixing layer before 3D patch embedding layer. The feature mixing helps in learning the inter-relationship between the features in terms of depth and channel. The output of the linear embedding of the patches forms the token provided to the transformer architecture. Output tokens are expanded and projected back to the original format. The model includes four encoder-decoder blocks, with each block on the encoder having four 3D transformer layers while we limit the number of layers on the decoder to one per block (encoder/decoder). We limited the number of blocks to four to better handle the data demanding nature of transformers as we do not pre-train our model on any other dataset \cite{Liu2021_swin_transformer, Liu2021_video_swin_transformer, Dosovitskiy2021_image_transformer, ranftl2021vision_densePrediction}. Likewise, we used \textit{mlp-ratio} of one in our architecture to reduce parameter count as opposed to the suggested value of four \cite{Liu2021_swin_transformer}. 
Transformer friendly augmentation including \textit{RandomHorizontalFlip} and \textit{RandomVerticalFlip} are best suited for the model training. These ensure that we can leverage data augmentation without making any change in the data presented except flipping. 

The building blocks of our model shown in fig. (\ref{fig: architecture}a) are detailed below.

\subsection{Feature Mixing Layer}
Mixing the time (t) and channel (c) dimensions of the spatiotemporal input (t x c x h x w) logically improves the interrelationship of window-based attention.  To achieve this, the feature is first transposed (reshaped) into h x w x (t x c), where the t and c dimension are flattened. The flattened dimension is then transformed using a fully connected layer. The transformed dimension is finally reshaped back into the previous spatiotemporal format. This operation improves performance of the model as shown in Table \ref{tab: model val result}.

\subsection{Patch Partitioning}
The 3D features (Spatiotemporal) are split into patches which are transformed to linear embedding. To achieve this, two layers of CNN starting with a strided convolution are used. The first convolution has a stride value equal to the dimension of its kernel size while the second has a kernel size of 1.

\subsection{Swin Transformer Block}
\label{sec: 3d swin tnx}
The transformer layer includes standard multi-head self attention (MSA), followed by a feed-forward network (MLP). Each of these layers is preceded by Layer Normalization (LN) in Vision transformer \cite{Dosovitskiy2021_image_transformer}, as opposed to post normalization used in NLP \cite{transformer_vaswani2017}. In this research, 3D shifted window MSA is employed, owing to the spatiotemporal nature of the input. The Swin transformer uses an interchange of sliding windows, as shown in Fig (\ref{fig: architecture}b and c) with a window (local) attention, followed by another local but shifted window attention. With such a setup, any two layers of attention follow (\ref{eqn: attn}):

\begin{equation}
    \label{eqn: attn}
    \begin{split}
    % \begin{align}
    \bar{z}^{l} = & \text{ W-MSA}(\text{LN}(z^{l-1})) + z^{l-1} \\
    z^{l} = & \text{ MLP}(\text{LN}(\bar{z}^{l})) + \bar{z}^{l} \\
    \bar{z}^{l+1} = & \text{ SW-MSA}(\text{LN}(z^{l})) + z^{l} \\
    z^{l+1} = & \text{ MLP}(\text{LN}(\bar{z}^{l+1})) + \bar{z}^{l+1}
    % \end{align}
    \end{split}
\end{equation}
where LN, MLP, W-MSA and SW-MSA represent layer normalization \cite{layer_norm}, multilayer perceptron, windowed multi-head self-attention and shifted window multi-head self-attention, respectively. The main difference between W-MSA and SW-MSA is the shift in window positioning, prior to computing local attention within the windowed blocks. Also, relative position bias is used in both W-MSA and SW-MSA \cite{Liu2021_swin_transformer, Liu2021_video_swin_transformer}. Following the work of \textit{Li et. al.}, we used a window size of (1, 8, 8), shift size of 2 in our implementation of the 3D Swin transformer block \cite{Liu2021_video_swin_transformer}. For the MLP, we used two fully-connected layers with a ratio of one for the hidden features (eqn. \ref{eqn: mlp}).

\begin{equation}
    \label{eqn: mlp}
    \text{MLP}(X) = (X W_1) W_2
\end{equation}
where $X \in \Re^{... \text{ x } d}$ is the input, $W_1 \in \Re^{d \text{ x } 4d}$ is the weight matrix of the first (hidden) fully-connected layer and $W_2 \in \Re^{4d \text{ x } d}$ is the weight matrix of the second (output) fully-connected layer.

\subsection{Patch merging layer}
This layer concatenates the features of each group of 2 × 2 neighboring patches, and applies a fully-connected layer on the 4C-dimensional concatenated features to get 2C-dimensional output \cite{Liu2021_swin_transformer, Liu2021_video_swin_transformer}. This results in a learned down-sampling operation.

\subsection{Patch expanding layer}
Contrary to the patch merging layer introduced in the Swin transformer \cite{Liu2021_swin_transformer, Liu2021_video_swin_transformer}, we used a fully connected layer to scale up the dimension of the incoming data. This results in a learned up-sampling operation.

\subsection{Encoder}
The encoder backbone network in our model includes a multi-stage Video Swin transformer. Specifically, we use four stages, each having four 3D transformer blocks, following a patch merging layer \cite{Liu2021_swin_transformer, Liu2021_video_swin_transformer}. The attention layer used here is the multi-head self attention as explained in section (\ref{sec: 3d swin tnx}) \cite{Liu2021_video_swin_transformer}.

\subsection{Decoder}
Here, we replaced the CNN blocks in the UNet with a 3D Swin-Transformers. Its structure is as shown in Fig (\ref{fig: architecture}a). The decoder starts with the upsampling of the incoming data using Patch expanding layer. We follow this with Feature Merging Layer which takes one of three possibilities
\begin{itemize}
    \item Concatenation: This uses simple concatenation similar to standard UNet \cite{unet}.
    \item Addition: This uses simple addition as used in LinkNet \cite{linknet_Chaurasia2017}.
    \item Both: combination of the above
\end{itemize}

\subsection{Neck}
As shown in Fig (\ref{fig: architecture}a), our proposed model links the encoder to the decoder via a \textit{neck} block.
% We use two Transformer blocks without a preceding patch merging layer as the \textit{neck} \cite{Cao2021_swin_unet}.
We limit the number of Transformer blocks to two in the \textit{neck} without a preceding patch merging layer. 
% Table (\ref{tab: model val result}) shows some models where we either use a neck or not in our experiments.

\subsection{Prediction Head}
This layer projects the output of the last decoder to the expected dimensions and format.
It comprises of a patch expanding layer to recover spatial dimension and also a fully connected layer to project the data into the desired dimensions (number of expected channels/frames).

\section{Experimental Results}
\label{sec: result}
% \subsection{Weather forecasting}
\paragraph{Data Description}: We used the dataset in the Traffic Map Movie Forecasting 2021 (Traffic4Cast2021) challenge for evaluation purposes \cite{traffic4cast2021}.
The challenge is to predict the dynamic traffic states 5, 10, 15, 30, 45 and 60 minutes into the future after 60 minutes time slots.
The dataset includes dynamic and static information covering 10 culturally diverse cities in a time span of 2 years and includes dynamic and static information.
The dynamic data are derived from GPS trajectories aggregated into spatiotemporal cells (495 x 436), with each cell corresponding to an area of approximately 100m x 100m and the time interval of 5 minutes.
The dynamic information is encoded in 8 channels containing the traffic volume and average speed per heading direction: NE, SE, SW, or NW. 
The static information describes the properties of the road maps and is split only into spatial cells. 
It is encoded in 9 channels representing the density of the road network and the road connections to the 8 neighboring cells. 
The traffic data can be presented as a movie with 288 frames per day for each city, thus effectively recasting traffic prediction as a video frame prediction task.
% Traffic4cast2021 is the third year of NeurIPS competition where participants are to predict traffic parameters (speed and volume) for four cardinal directions (NorthEast, SouthEast, NorthWest and SouthWest).
This competition has two challenges:
\begin{itemize}
    \item Core Challenge: data contain the training and test sets for four cities \{Berlin, Chicago, Istanbul and Melbourne\}. 
    \item Extended Challenge: data contain only the test set for two additional cities \{New York and Vienna\}
\end{itemize}
The dataset also includes four additional cities that can be used for pretraining.

\paragraph{Model Training}:
The model described in Fig. \ref{fig: architecture} was implemented in Pytorch. % \cite{NEURIPS2019_9015_pytorch}. 
The mean squared error (MSE) was used as the loss function, with the Adam optimizer \cite{Kingma2015AdamAM}. 
The learning rate was initially set to 1e-4 and was manually reduced to 1e-7, when performance plateaued on the (randomly selected) validation set. 
% Our model was trained with carefully considered data augmentation for segmentation purposes. Specifically, we use \textit{RandomHorizontalFlip} and \textit{RandomVerticalFlip}, which ensures that we can leverage data augmentation without making any change in the data presented except flipping. 
% This is important as we train a single model with data from the three provided regions (R1, R2, R3), while we tested the model on all available regions (R1-R6) to account for the transfer learning challenge.

% As shown in Table (\ref{tab: results compared}), the proposed model resulted in an MSE of \textit{0.5337} and \textit{0.4959}, for the \textit{core} and \textit{transfer} challenges, respectively, with only \textit{688,080} parameters. 
% The proposed model was placed in the fourth position on the final leaderboard, while having a smaller number of parameters. 
% Moreover, this is the first instance of using vision transformers in spatiotemporal forecasting. 
As shown in Table (\ref{tab: model val result})
, we trained three different model  configurations with an embedding dimension of {96, 192}, with or without Feature Mixing layer. 
% We considered the use of a \textit{neck} in our \textit{UNet-like} model resulting in a conclusion that this does not help our model to improve (Table \ref{tab: model val result}). 
In Table (\ref{tab: results compared}) , we compared our model with the baseline models (\textit{GCN} and \textit{UNet}) provided on the leaderboard reflect our model's performance. We would like to note that score computation on the leaderboard was done without data normalization.
% In Tables (\ref{tab: model val result}), % \& \ref{tab: model result}),
% it is important to note that the \textit{validation} MSE  is computed on the validation set (Region R1-R6 \cite{traffic4cast2021}) 
% while the \textit{core} and \textit{transfer} MSE 
% in Table (\ref{tab: results compared}) 
% are computed using a specially crafted MSE computation which puts a segmentation mask into consideration \cite{traffic4cast2021}.

\begin{table}[htbp]
  \caption{Results of our various model configurations on the Traffic4Cast2021 Data}
  \label{tab: model val result}
  \centering
  \begin{tabular}{ll|r|l}
    \toprule
    dim & Mix Features & \#Parameters  & Core MSE\\
    \midrule
    96 & False & 8,935,246 & 50.2393\\
    192 & False & 141,820,294 & 49.9917\\
    \textcolor{blue}{192} & \textcolor{blue}{True} & \textcolor{blue}{141,848,694} & \textcolor{blue}{49.7208}\\
    \bottomrule
  \end{tabular}
\end{table}

\begin{table}[htbp]
  \caption{Comparing our result with baseline models' results on Traffic4Cast2021 Data}
  \label{tab: results compared}
  \centering
  \begin{tabular}{lll}
    \toprule
    Method & Core MSE \\
    \midrule
    GCN baseline & 51.7143\\
    Unet-baseline & 51.2826\\
    \textit{Swin-UNet3D(Ours)}	    &   \textcolor{blue}{49.7208}\\
    \bottomrule
  \end{tabular}
\end{table}

\section{Conclusions and Future Work}
We presented the use of 3D Swin-Transformer in a UNet architecture for short time spatiotemporal forecasting, which resulted in competitive results, i.e., an MSE of \textit{49.7208} for the \textit{core} challenges (Traffic4Cast2021 \cite{traffic4cast2021}). 
% The proposed model was placed in the fourth position on the final leaderboard, while having a smaller number of parameters. 
The model having only four blocks in Swin-transformers in both encoder and decoder was implemented in PyTorch and trained using \textit{Pytorch-Ligthning} \cite{falcon2019pytorch}. In the future, we plan to use multi-objective loss formulation for the multi-task model training \cite{NeurIPS2018_Sener_Koltun_multi_objective}.
As transformer architectures are still relatively new to the computer vision domain, we plan to explore other variants of attention layers in the future. Likewise, we equally plan to explore token mixing using hypercomplex networks \cite{bojesomo2020traffic}.
Source code with the implementation of the proposed approach is available online at \href{href: code}{\textit{https://github.com/bojesomo/Traffic4Cast2021-SwinUNet3D}}.

\section*{Acknowledgments}
  This work was supported by the ICT Fund, Telecommunications Regulatory Authority (TRA), Abu Dhabi, United Arab Emirates.

%%
%% Define the bibliography file to be used
\bibliographystyle{IEEEtran}
\bibliography{references}


\end{document}